\documentclass[letterpaper, 10 pt, conference]{main} 
                                                        
\IEEEoverridecommandlockouts                             
                                                         
\usepackage{graphics} 
\usepackage{epsfig} 
\usepackage{mathptmx} 
\usepackage{amsmath} 
\usepackage{amssymb}  
\usepackage{xcolor}
\usepackage{multirow}
\usepackage{tabularx}
\usepackage{booktabs}
\usepackage{multirow}
\usepackage{booktabs,ragged2e}
\usepackage{hhline}
\usepackage{cellspace} 
\usepackage{subcaption}
\usepackage{hyperref}
\usepackage{algorithmic}
\usepackage[inkscapelatex=false]{svg}
\usepackage{graphicx}
\usepackage{array} 
\usepackage{placeins}
\usepackage{comment}

\title{\LARGE \bf
Large Language Model-based Decision-making for COLREGs and the Control of Autonomous Surface Vehicles}

\author{Klinsmann Agyei$^\dag$, Pouria Sarhadi$^\dag$, Wasif Naeem$^\ddag$
\thanks{$^\dag$Klinsmann Agyei and Pouria Sarhadi are with the School of Physics, Engineering and Computer Science, University of Hertfordshire, Hatfield, United Kingdom        ({\tt\small p.sarhadi@herts.ac.uk)}}%
\thanks{$^\ddag$Wasif Naeem is with the School of Electronics, Electrical Engineering and Computer Science
Queen’s University Belfast, Belfast, UK ({\tt\small w.naeem@qub.ac.uk)}}%
\thanks{©2025 IEEE.  Personal use of this material is permitted.  Permission from IEEE must be obtained for all other uses, in any current or future media, including reprinting/republishing this material for advertising or promotional purposes, creating new collective works, for resale or redistribution to servers or lists, or reuse of any copyrighted component of this work in other works.}
}

\begin{document}

\maketitle
\thispagestyle{empty}
\pagestyle{empty}

\begin{abstract}
In the field of autonomous surface vehicles (ASVs), devising decision-making and obstacle avoidance solutions that address maritime COLREGs (Collision Regulations), primarily defined for human operators, has long been a pressing challenge. Recent advancements in explainable Artificial Intelligence (AI) and machine learning have shown promise in enabling human-like decision-making. Notably, significant developments have occurred in the application of Large Language Models (LLMs) to the decision-making of complex systems, such as self-driving cars. The textual and somewhat ambiguous nature of COLREGs (from an algorithmic perspective), however, poses challenges that align well with the capabilities of LLMs, suggesting that LLMs may become increasingly suitable for this application soon. This paper presents and demonstrates the first application of LLM-based decision-making and control for ASVs. The proposed method establishes a high-level decision-maker that uses online collision risk indices and key measurements to make decisions for safe manoeuvres. A tailored design and runtime structure is developed to support training and real-time action generation on a realistic ASV model. Local planning and control algorithms are integrated to execute the commands for waypoint following and collision avoidance at a lower level. To the authors' knowledge, this study represents the first attempt to apply explainable AI to the dynamic control problem of maritime systems recognising the COLREGs rules, opening new avenues for research in this challenging area. Results obtained across multiple test scenarios demonstrate the system's ability to maintain online COLREGs compliance, accurate waypoint tracking, and feasible control, while providing human-interpretable reasoning for each decision.
\end{abstract}

\section{INTRODUCTION}
Rapid advancements in Artificial Intelligence (AI) over recent years have led to breakthroughs in numerous applications, including autonomous vehicles. Generative AI and Large Language Models (LLMs), with capabilities in interpreting text, voice, and video, show promise across many systems \cite{cui2024survey, chen2024llm, chen2024driving}. Meanwhile, the maritime vehicle systems community has shown interest in Machine Learning (ML) solutions \cite{sarhadi2022survey}. These technologies are expected to improve efficiency, safety, and cost-effectiveness, though their deployment still requires further exploration. This situation is comparable with the early promotions of automatic control in the early and mid 20th century, which gained wider accessibility later. To date, safe ASV decision-making has relied primarily on rule-based approaches with performance trade-offs, while high-performance solutions are actively being investigated \cite{sarhadi2022survey, maza2022colregs, vagale2021review}. LLMs' ability to process natural language, contextualisation, and generate human-like reasoning has opened new possibilities for autonomous decision-making systems \cite{cui2024survey, chen2024llm, chen2024driving,lin2024drplanner}. COLREGs-compliant decision-making in the maritime environment is one of the pressing challenges for ASVs, where textual, non-algorithmic rules need to be translated into machine language. An early consideration in the COLREGs3 project \cite{Sanchez-Heres_et_al_2023} highlighted potential benefits, with some scene-based tests but without dynamical simulation, resulting in poor outcomes and identifying a need for re-training LLMs. Similarly, another study on ship collision avoidance suggested that LLMs could introduce a paradigm shift in ship decision-making \cite{wang2024multipleship}. This paper develops a new framework for LLM-based, risk-aware decision-making integrated with low-level planning and control algorithms, applied to a nonlinear ASV model. We demonstrate the first instance of online, explainable decision-making and control across various encounter scenarios.

The integration of LLMs into autonomous systems has seen growing interest across various domains. Chen et al. \cite{chen2024llm} introduced an LLM-driven framework for multiple-vehicle dispatching, demonstrating the model's capability to process real-time traffic data and environmental conditions for optimal routing in urban environments. Their work showed how LLMs can effectively handle complex traffic scenarios while maintaining safety constraints. A comprehensive survey \cite{chen2024driving} explored multimodal LLMs in autonomous driving, highlighting the challenges and opportunities in integrating diverse sensory information for robust decision-making. This integration of multiple data modalities with LLM reasoning capabilities has proven particularly effective in dynamic environments. Building on these foundations, several groundbreaking approaches have emerged. Significant advances in traffic management were achieved through human-mimetic traffic signal control \cite{wang2024llm}, while \cite{li2024driving} focused on policy adaptation across diverse driving environments, demonstrating LLMs' flexibility in handling varying conditions. DILU \cite{wen2023dilu} presented a knowledge-driven approach incorporating human like reasoning, addressing one of the key challenges in autonomous systems, the integration of human-like understanding with machine precision. Another research is LanguageMPC \cite{sha2023languagempc}, which showed how LLMs can enhance autonomous systems' ability to interpret complex rules while maintaining safety and efficiency. This work demonstrated the potential for LLMs to bridge the gap between regulations written for human consumption and machine execution. In \cite{fu2024drive}, the authors presented a `Drive like a Human' approach by demonstrating LLMs' capability to mirror human driver behaviour through natural language reasoning, achieving a level of efficiency in decision-making previously difficult to attain with traditional approaches.

\section{Problem statement: Mission planning and COLREGs-based decision-making}
Maritime navigation is governed by the International Regulations for Preventing Collisions at Sea or COLREGs, a complex set of rules that define how vessels should behave in various encounter situations \cite{imo_colregs}. These rules, particularly Rules 13-17, establish clear protocols for overtaking, head-on, and crossing situations, requiring vessels to make predictable and consistent decisions \cite{johansen2016ship}. However, interpreting and applying these rules in real-time autonomous navigation presents significant challenges due to the dynamic nature of maritime encounters and the need for human-like reasoning in situation assessment.

Recent advancements in LLMs have demonstrated remarkable success in complex decision-making tasks, particularly in autonomous driving applications \cite{chen2024llm,chen2024driving,lin2024drplanner,wang2024llm,li2024driving,atakishiyev2024explainablellmreview} where they enable human-like reasoning and clear explanation of actions. Studies have shown LLMs' capability to interpret traffic rules, assess dynamic situations, and generate appropriate responses in road environments. While autonomous driving have seen substantial integration of LLMs in their decision-making algorithms , with numerous successful implementations and ongoing research, the application to maritime collision regulations (COLREGs) compliance remains largely unexplored. Also, The absence of human-like decision-making capabilities in ASVs poses significant safety risks as the industry moves toward higher levels of automation. Traditional rule-based systems often struggle with the nuanced interpretation of COLREGs rules and fail to provide clear explanations for their actions, limiting trust and adoption in practical applications. There is a critical need for systems that can combine sophisticated situational awareness with clear, explainable decision-making that mirrors trained mariners expertise.

\begin{figure}[t]
    \centering    \centerline{\includegraphics[width=0.5\textwidth]{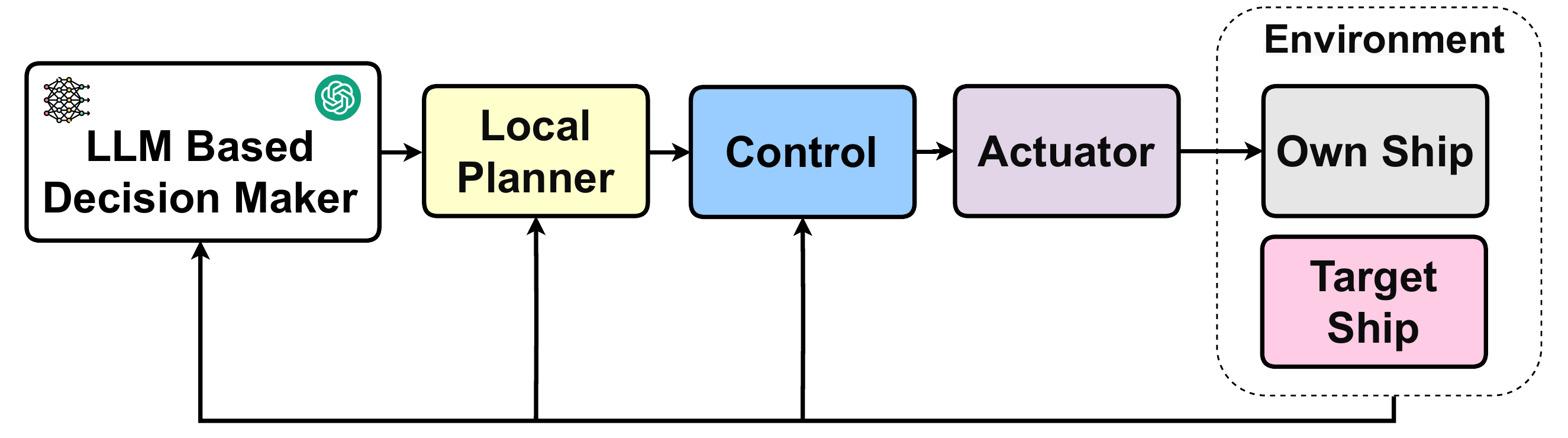}}
    \caption{The proposed autonomy algorithm block-diagram}
    \label{fig:proposed solution}
\end{figure}

This research addresses these challenges through a novel architecture (shown in Fig.~\ref{fig:proposed solution}) that integrates LLM-based decision-making with local planning and control execution for ASV navigation. The proposed solution bridges the gap between high-level COLREGs interpretation and practical vessel control, enabling consistent and explainable collision avoidance behaviours.

\section{The Proposed LLM-based risk-aware decision-making and control method}
The proposed scheme develops a high-level LLM-based risk-aware decision-making algorithm as shown in Fig. \ref{fig:System Architecture} and discussed in this Section.

\subsection{LLM-based decision-making algorithm}
\begin{figure*}[t]
    \centering    \centerline{\includegraphics[width=0.98\textwidth]{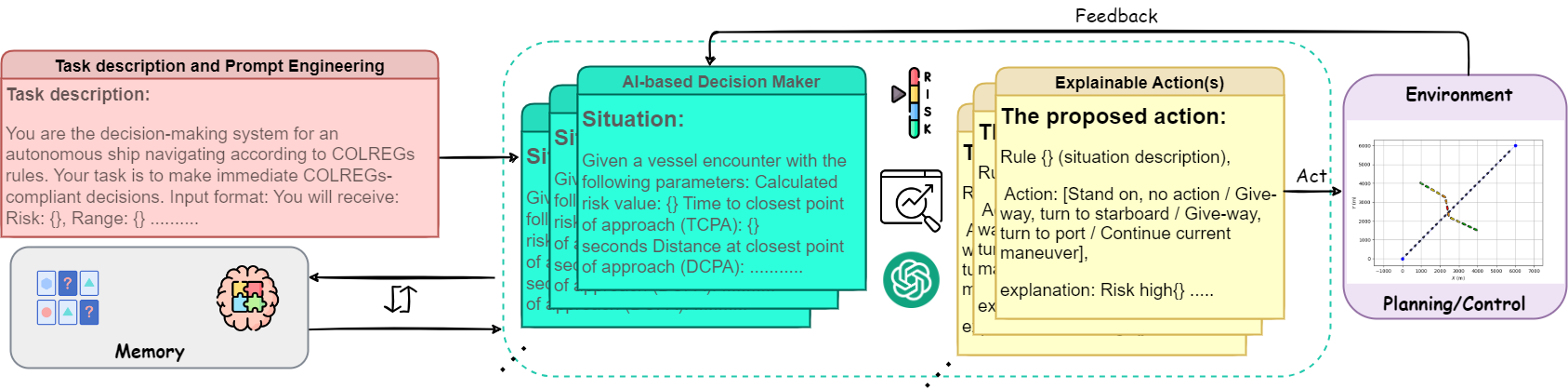}}
    \caption{The algorithm structure for design- and run-time training of the LLM-based decision maker incorporating risk}
    \label{fig:System Architecture}
\end{figure*}

The proposed COLREGs-compliant decision-making system employs a novel architecture that leverages LLMs for maritime decision-making. Our system utilised OpenAI's GPT-4 model accessed through the ChatGPT API, with a temperature setting of 0.2 to ensure consistent and deterministic outputs. The LLM was not specifically trained on maritime data; instead, we leveraged its pre-trained and general reasoning capabilities through careful prompt engineering that explicitly encoded COLREGs rules and maritime navigation requirements. The complete implementation is available online via GitHub\footnote{\href{https://github.com/Klins101/CORALL}{\textcolor{blue}{github.com/Klins101/CORALL}}}, including all Python code, prompt templates, and simulation environments used in this study. The architecture consists of four primary components that work in concert to generate safe and explainable navigation decisions as illustrated in Fig.~\ref{fig:System Architecture}. At the highest level, the Task Description and Prompt Engineering component establishes the fundamental framework for the LLM's operation. This component transforms maritime navigation expertise into a structured prompt that guides the LLM's understanding of COLREGs rules, relative heading interpretations, and decision-making requirements.

The core of our architecture lies in the LLM-based Decision Maker, which processes current encounter conditions to generate appropriate actions. This component evaluates a specific set of parameters: the relative heading ($\psi_{rel}$) between own vessel and target vessel, risk assessment (Risk) calculated through fuzzy logic, Distance to Closest Point of Approach ($D_{CPA}$), Range, and Time to Closest Point of Approach ($T_{CPA}$). Additionally, it considers the current manoeuvre status (give-way/stand-on) to maintain decision consistency. By framing these parameters as a condition-based query, the LLM determines the appropriate COLREGs-compliant action for the current situation. The Explainable Action component ensures that all decisions are transparent and interpretable. It generates standardised outputs that include situation classification (overtaking, head-on, or crossing), specific action commands (stand-on or give-way), and clear reasoning chains explaining the decision basis as illustrated in Fig.~\ref{fig:Prompt}. 
Supporting these primary components, the manoeuvre or memory block retains the encounter state and ensures decision consistency throughout manoeuvres. It stores crucial information about the initial situation classification, current action status, and navigation history. This state of awareness prevents oscillatory behaviour and ensures smooth and predictable decision-making. The Maritime Environment component provides continuous feedback about encounter geometry and risk evolution, enabling the system to adapt to changing conditions while maintaining COLREGs compliance.

Our solution addresses three key considerations:
\subsubsection{Consistency of Situational Awareness} 
The maritime environment requires precise interpretation of relative headings for accurate scenario classification. The system introduces a rigorous classification framework as follows:

A structured semantic interpretation of relative heading $\psi _{rel}$ is employed between own ship (OS) and target ship (TS) through a formalised prompt engineering approach. The situation classification function $S(\psi _{rel})$ is implemented through an LLM with a domain-specific maritime navigation template $\mathcal{M}$:
\begin{equation}
\mathcal{M} = \{ \mathcal{B}, \mathcal{R}, \mathcal{D} \} 
\end{equation}
where:
\begin{itemize}
    \item $\mathcal{B}$: Bearing interpretation mapping
    \item $\mathcal{R}$: Rule-specific constraints
    \item $\mathcal{D}$: Decision parameters
\end{itemize}
The bearing interpretation mapping $B$ is defined as:
\begin{equation} 
\footnotesize
    \mathcal{B}(\Psi_{rel}) = 
    \begin{cases} 
        \text{Head-on}, & \text{if } -6^\circ \leq \Psi_{rel} \leq 6^\circ \\
        \text{Overtaking}, & \text{if } -112^\circ \leq \Psi_{rel} \leq 112^\circ \\
        \text{Crossing}, & \text{otherwise}
    \end{cases}
\end{equation}

This mapping is encoded in the LLM's prompt architecture through a structured maritime domain language that ensures consistent interpretation across all encounter scenarios.

\subsubsection{Risk Assessment Integration}

The system incorporates a real-time risk evaluation function $Risk(t)$ as detailed in \ref{Rule-based local planning and control algorithms} with critical threshold vector $\mathcal{T}$:
\begin{equation}
\footnotesize
\mathcal{T} = 
\begin{cases}
   \mathcal{T}_{Risk}: 0.75, \\
   \mathcal{T}_R: 1000\ \text{m}, \\
   \mathcal{T}_{D_{CPA}}: 250\ \text{m} \\
   \mathcal{T}_{T_{CPA}}: 60\ \text{s}
\end{cases}
\end{equation}

\subsubsection{State-Aware Decision Maintenance}

The system maintains the encounter state through a state vector $S$:
\begin{equation}
S = \{\sigma, \alpha, \gamma, i\}
\end{equation}
\vspace{-5pt}
where:
\begin{itemize}
    \item $\sigma \in \Sigma$: situation space \{overtaking, head-on, crossing\}
    \item $\alpha \in A$: action space \{stand-on, give-way\}
    \item $\gamma \in T$: turning state \{true, false\}
    \item $i \in \mathbb{N}$: manoeuvre initiation index
\end{itemize}

The complete decision-making process can be expressed as:
\begin{equation}
D(\psi_{rel}, S) = LLM(\mathcal{M}, B(\psi_{rel}), Risk, S)
\end{equation}

where $D$ represents the final decision function that maps the input space to the action space while maintaining COLREGs compliance.

\subsection{The environment and low-level algorithms' simulation} \label{Rule-based local planning and control algorithms}
The environment simulation includes the vehicle model, low-level local planning, control algorithm, and risk assessment, as discussed in \cite{UKACC2022}. A brief explanation of the process is provided below. The Python code used to simulate the system is available online via GitHub\footnote{\href{https://github.com/Psarhadi/Autonomous_ship_planning_collision_avoidance_control}{\textcolor{blue}{github.com/Psarhadi/Autonomous\_ship\_planning\_control}}}.
\subsubsection{Vehicle model}
A nonlinear model of a marine vessel is considered as follows:
\begin{equation}
\left\{
\begin{aligned}
\dot{X}(t) &= u(t) \cos(\psi(t)) \\
\dot{Y}(t) &= u(t) \sin(\psi(t)) \\
\dot{\psi}(t) &= r(t) \\
\dot{r}(t) &= -(1/T_\psi) r(t) + (K_\psi/T_\psi)u_c(t) \\ 
\dot{u}(t) &= -(1/T_u) u(t) + (K_u/T_u) \big(\tau(t) - d(t)\big) \\
\dot{d}(t) &= -(1/T_d) d(t) + \omega_d(t) 
\end{aligned}
\right.
\label{eq:shipmodel}
\end{equation}
In Eq. \ref{eq:shipmodel}, $[X,Y,\psi]$ represents the vehicle pose, including longitudinal and lateral positions and the yaw angle. The variable $r$ denotes the yaw rate with Nomoto's first-order dynamics, with $T_\psi$ and $K_\psi$ as the time constant and gain. The control torque $|u_c|<\pm20$ is the input with amplitude saturation. A first-order transfer function models the vessel speed $u$, with $T_u$ and $K_u$ as parameters. The variable $\tau$ is the input, while $d$ is a disturbance modelled by a Markov process. A random Gaussian signal $\omega_d$ is the disturbance input, and $T_d$ is its time constant. This model captures the main characteristics of a ship and its manoeuvring constraints, providing a semi-realistic test environment for training and testing the algorithms.
\subsubsection{Local planning algorithm}
\begin{figure}[b]
    \centering
    \centerline{\includegraphics[width=0.42\textwidth]{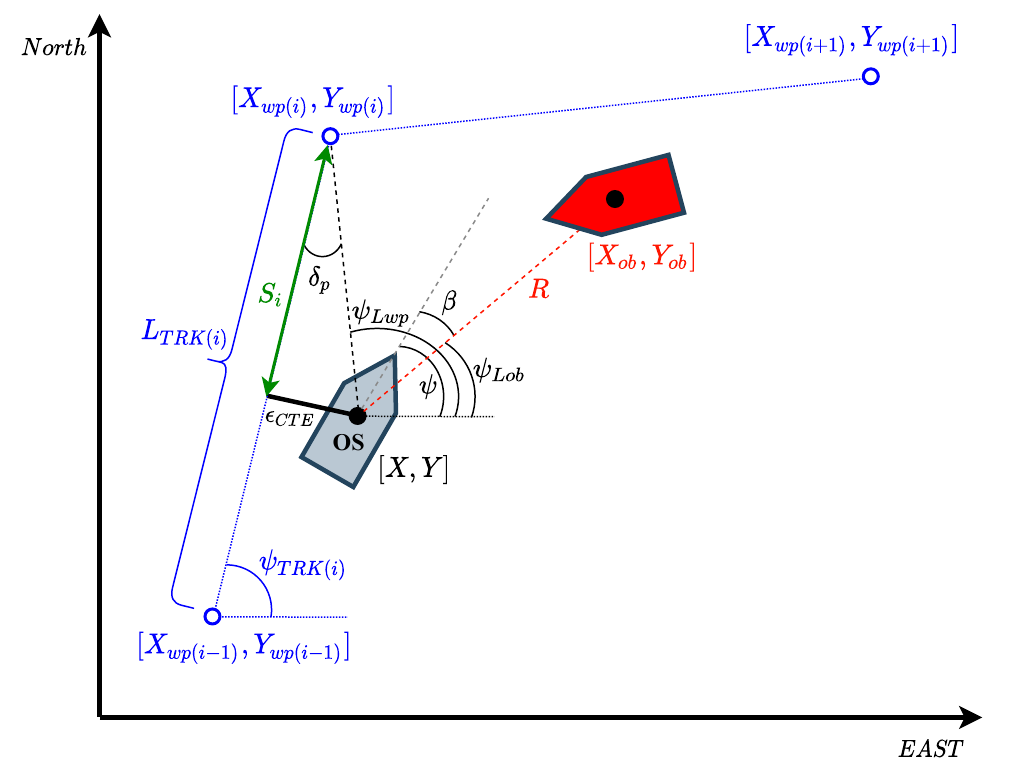}}
    \vspace{-10pt}
    \caption{Waypoint following scene and its parameters}
    \label{fig:Planning}
\end{figure}

The proposed local planning algorithm generates three main terms to follow consecutive waypoints while avoiding obstacles, as shown in \cite{UKACC2022}:
\begin{equation}
    \psi_d(t) = \psi_{LOS}(t) + \psi_{CTE}(t) + \psi_{COLAV}(t)
    \label{eq:psid}
\end{equation}
Eq. \ref{eq:psid} consists of a Line of Sight (LOS) term, a Cross-Tracking Error (CTE) term, and Collision Avoidance (COLAV) inputs to follow consecutive $X$ and $Y$ waypoint sets: $WP(i) = \begin{bmatrix} X_{wp(i)}, Y_{wp(i)} \end{bmatrix}$. A geometrical illustration is shown in Fig.~\ref{fig:Planning}.
The LOS tracking term, $\psi_{LOS}$, is defined as:
\begin{equation}
\psi_{LOS}(t) = \text{atan2}(Y_{ewp(i)}(t), X_{ewp(i)}(t))
\end{equation}
where $X_{ewp(i)}(t)$ and $Y_{ewp(i)}$ are the tracking errors in the $X$ and $Y$ directions, i.e., $X_{ewp(i)}(t) = X_{wp(i)} - X(t)$ and $Y_{ewp(i)}(t) = Y_{wp(i)} - Y(t)$. To minimise the cross-tracking error ($\epsilon_{CTE}$ in Fig. \ref{fig:Planning}), the following expression is used:
\begin{equation}
\psi_{LOS}(t) = - \text{atan}\left(\epsilon_{CTE}(t)/\mu \right)
\end{equation}
where $\mu$ is a tuning parameter and $\epsilon_{CTE}$ is calculated as:
\begin{equation}
\epsilon_{CTE}(t) = S_i(t) \tan\left( \delta_p(t) \right)
\end{equation}
In the above, $\delta_p(t)$ represents the angle to the next waypoint, and $S_i$ is the along-track error, as shown in Fig. \ref{fig:Planning} They can be calculated using the geometry of the encounter (more details in \cite{UKACC2022}). The COLAV term for the heading is generated based on a reactive avoidance approach in the following form:
\begin{equation}
\psi_{COLAV} = K_{Dir} K_{COLAV} w_{R} w_{\beta}
\end{equation}
This equation assigns two main weights, $w_{R}$ and $w_{\beta}$, for the distance ($R$) and bearing ($\beta$) angle to the obstacle. The parameter $K_{COLAV}$ is a tuning gain for the COLAV algorithm, and $K_{Dir}$ defines the desired action from the decision-making algorithm. Together, these three terms in Eq.~\ref{eq:psid} establish path following and COLAV to imitate the ship's behaviour.
\subsubsection{Control algorithm}
The controller consists of a Proportional-Derivative (PD) controller, with gains ($K_p$ and $K_d$) that can be gain-scheduled for different ASV speeds. It takes the following form to generate the control actions ($u_c$ in Eg. \ref{eq:shipmodel}) to follow the desired heading:
\begin{equation}
u_c(t) = K_p (\psi_{d}(t) - \psi(t)) - K_d r(t)
\end{equation}
An amplitude saturation is applied, as explained in the modelling section, to account for the system's manoeuvrability limitations.
\subsubsection{Online collision risk-assessment}
Another crucial item is risk assessment, which serves as one of the key inputs to the decision-making algorithm. It uses CPA analysis with its core indicators: distance and time to CPA ($D_{CPA}(t)$, $T_{CPA}(t)$), calculated using the following formulae:
\begin{equation}
D_{CPA}(t) = R(t) \sin(\alpha(t))
\end{equation}
\begin{equation}
T_{CPA}(t) = R(t) \cos(\alpha(t))/V_{rel}(t)
\end{equation}
where $V_{rel}$ is the relative velocity between the TS and OS, and $\alpha(t)$ is the corresponding angle. A method for calculating these parameters is provided in \cite{UKACC2022}. The risk index ($Risk$) at each iteration is calculated by Eq.~\ref{eq:Risk}:
\begin{equation}
Risk(t) = (f(D_{CPA}) + f(T_{CPA}) + f(R(t)))/3
\label{eq:Risk}
\end{equation}
where the function $f(.)$ is a Z-shaped Fuzzy Membership (ZMF) function, with two thresholds defining a weight that increases from 0 to 1 as the input decreases. Thus, the risk index is always a normalised value between 0 and 1, quantifying the risk and used in the LLM-based decision-making algorithm.

\section{Simulation Results}
\begin{figure*}[t]
    \centering
    \begin{subfigure}[b]{0.38\textwidth}
        \centering
        \includegraphics[width=\textwidth]{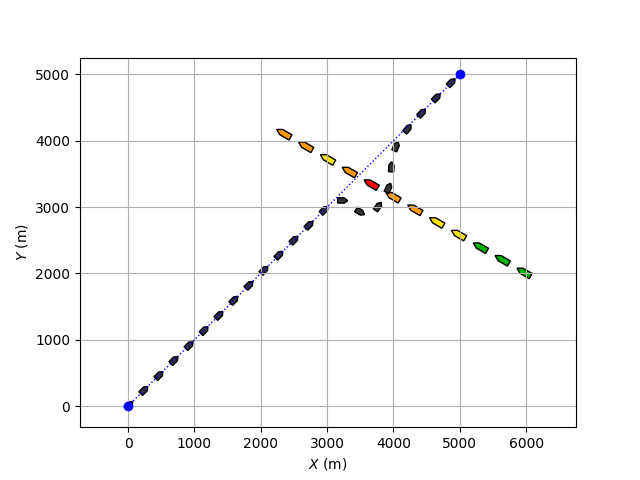}
        \caption{\scriptsize Crossing give-way encounter}
        \label{fig:Crossinggiveway}
    \end{subfigure}%
    \begin{subfigure}[b]{0.38\textwidth}
        \centering
        \includegraphics[width=\textwidth]{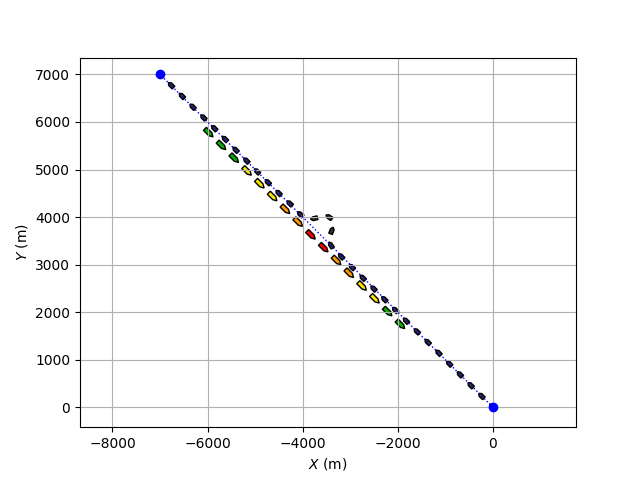}
        \caption{\scriptsize Head-on move to starboard encounter}
        \label{fig:headon}
    \end{subfigure}
    \vspace{0pt}
    
    \begin{subfigure}[b]{0.38\textwidth}
        \centering 
        \includegraphics[width=\textwidth]{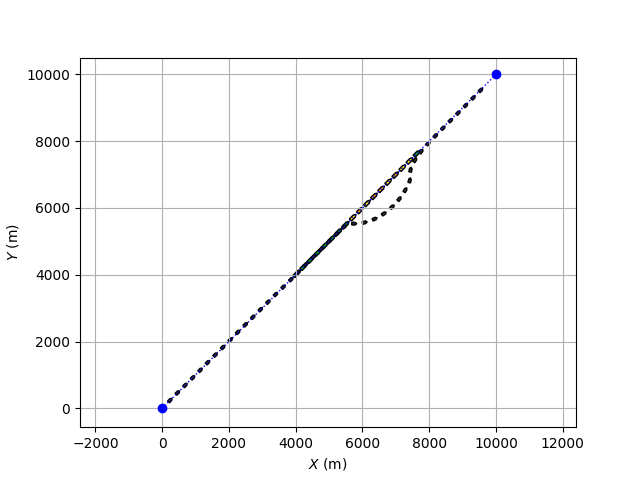}
        \caption{\scriptsize Overtaking encounter}
        \label{fig:overtaking}
    \end{subfigure}
    \begin{subfigure}[b]{0.38\textwidth}
        \centering
        \includegraphics[width=\textwidth]{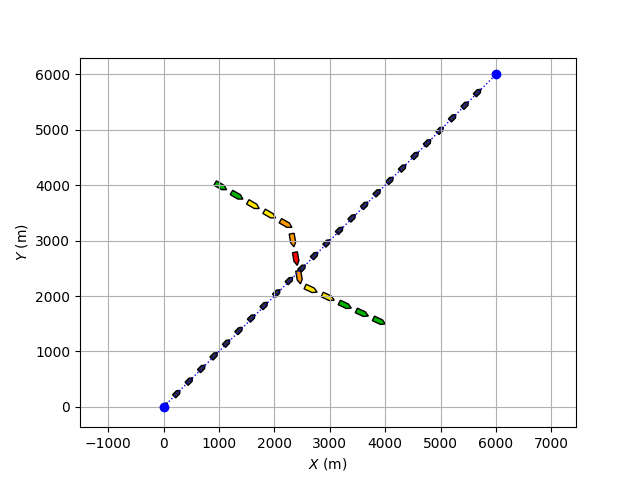}
        \caption{\scriptsize Crossing stand-on encounter}
        \label{fig:Crossingstandon}
    \end{subfigure}
    \vspace{-0pt}
    \caption{Illustration of the ASV motion in different scenarios.}
    \label{fig:ASV4SimulationScenes}
\end{figure*}

In this section, we evaluate the performance of the proposed LLM-based algorithm based on risk awareness in four typical scenarios of maritime encounters. The simulations were conducted at a high frequency, with the decision-maker operating at 1~Hz in an online environment. It should be noted that considering the dynamics of the ship, lower-level control loops run at 100~Hz. However, a frequency of 1~Hz is considered feasible for higher-level decision-making processes. In all simulations, the own ship maintained a constant cruising speed of 32 knots (16 m/s), chosen to reflect the operational challenges of a high-speed vessel, with possible collision. The traversed paths for the ASV (black) and the target ship (coloured based on the risk) for all cases are depicted in Fig.~\ref{fig:ASV4SimulationScenes}. In the crossing give-way scenario, the system demonstrated appropriate Rule 16 compliance by initiating a timely starboard turn when collision risk became significant. For head-on encounters, the system exhibited correct interpretation of Rule 14 requirements. When presented with critical parameters indicating an imminent head-on situation, the LLM generated clear decisions for starboard turn manoeuvres, ensuring mutual action compliance as required by COLREGs and as shown in Fig.~\ref{fig:ASV4SimulationScenes}. The overtaking scenario highlighted our system's capacity for sustained decision consistency. The LLM maintained appropriate Rule 13 compliance throughout the extended manoeuvre, as evidenced by its explicit reasoning about overtaking responsibilities and required actions as specified above. 
\vspace{0pt}

\begin{figure}[b]
    \centering    \centerline{\includegraphics[width=0.5\textwidth]{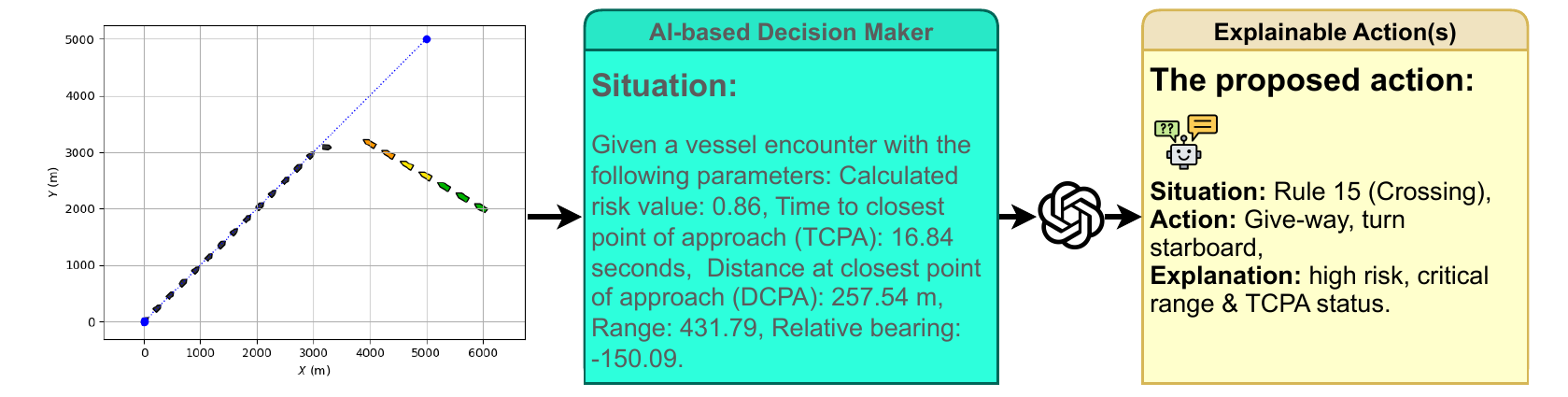}}
    \caption{decision-making, crossing give-way}
    \label{fig:Prompt}
\end{figure}

Notably, In the crossing stand-on situation, the algorithm effectively implemented Rule 17. Despite moderate-risk inputs, the LLM maintained correct stand-on behaviour with the target on the port side, providing clear explanations based on Rule 17. With the proposed strategy (Fig. \ref{fig:System Architecture}), the algorithm is able to make online, in-the-loop decisions throughout testing.
We further discuss the results of our prompt engineering for two critical situations. \\
Fig.~\ref{fig:Prompt} illustrates a moment in the crossing scenario, demonstrating the LLM's decision-making process in a critical situation. The trajectory plot (left) shows a developing crossing scenario with converging vessels. The algorithm processes a set of urgent parameters: high risk value of 0.86, critically low $T_{\text{CPA}}$ of 16.84~s, $D_{\text{CPA}}$ of 257.54~m, range 431.79~m, and relative heading of -150.09 $^{\circ}$. Through the prompt engineering framework (centre), the LLM correctly identifies this as a crossing situation where the own vessel is the give-way vessel. The relative heading of -150.09$^{\circ}$ (normalised to 209.91$^{\circ}$) places the target vessel on our starboard side, making the own vessel the give-way vessel under Rule 15. The Explainable Action(s) panel (right) shows the LLM's clear reasoning process, resulting in the decision to `Give-way, turn starboard'. This action is mandated by both our give-way obligation under Rule 15 and the critical nature of the situation, as evidenced by the high risk value, close range, and critically low $T_{CPA}$. This scenario verifies the system's ability to correctly interpret relative bearings for proper COLREGs classification while also considering the urgency of the situation in its decision-making process. 

\begin{figure}[t]
    \centering    \centerline{\includegraphics[width=0.5\textwidth]{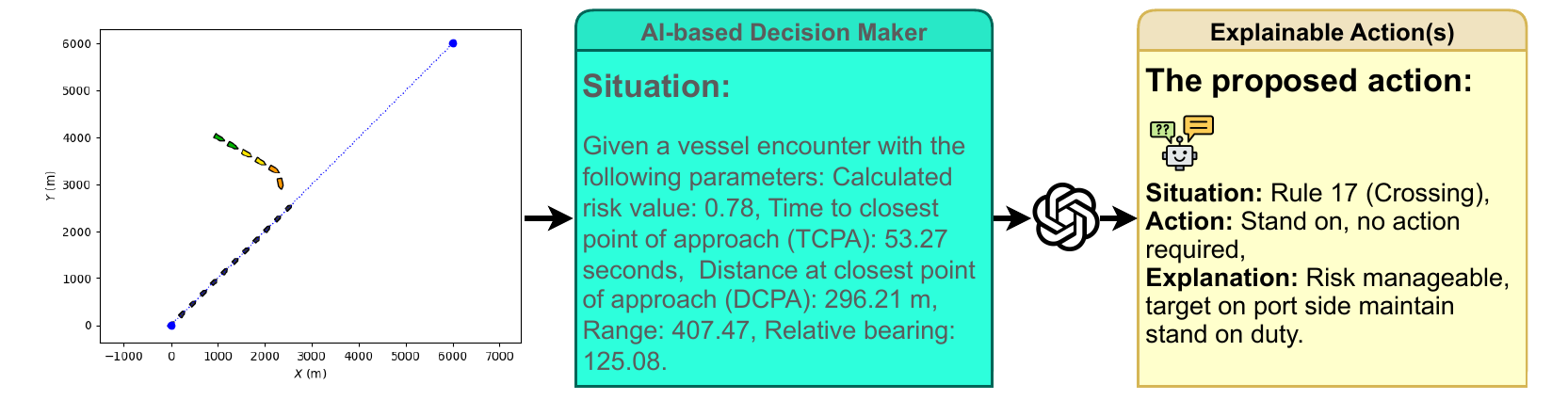}}
    \caption{decision-making, crossing stand-on}
    \label{fig:Prompt2}
\end{figure}

\vspace{-0.5pt}
For the crossing stand-on scenario illustrated in Fig.~\ref{fig:Prompt2}, we observe another demonstration of the LLM's decision-making capability. The trajectory plot (left) shows the evolution of a crossing situation where the target vessel approaches from the port side. The LLM-based decision maker processes a set of parameters indicating a manageable situation: calculated risk value of 0.78, $T_{CPA}$ 53.27s, $D_{CPA}$ 296.21m, range 407.47m, and relative heading 125.08$^\circ$. When this data is processed through our prompt engineering framework (centre), the LLM demonstrates precise understanding of COLREGs Rule 17 requirements for stand-on vessels. The Explainable Action(s) panel (right) shows the system's clear decision-making logic. With a moderate risk value (0.78) and parameters near but not exceeding critical thresholds, the LLM correctly maintains stand-on behaviour as required when the target vessel is on the port side. The explanation specifically acknowledges 'Risk present but manageable, maintaining stand-on duty as target vessel on the port side,' demonstrating the system's ability to balance risk awareness with proper COLREGs compliance. This scenario particularly highlights our system's sophisticated state awareness capabilities. Even with elevated risk values, the LLM maintains appropriate stand-on behaviour while monitoring the situation, precisely as required by Rule 17.

This diversity of appropriate responses across different encounter types, all generated with clear reasoning chains, verifies our LLM's capability to interpret and respond to various maritime encounters while maintaining regulatory compliance.

\section{CONCLUSION}
This paper presented a novel method to maritime COLAV for ASVs using an LLM-based algorithm for COLREGs-compliant decision-making. Our approach utilised a combination of a high-level, risk-aware LLM-based decision maker and rule-based low-level algorithms to perform path tracking with safe manoeuvres in collision encounters. This algorithm generates human-like, explainable interpretations in complex scenarios, proposing appropriate actions to achieve the mission objectives. The devised decision-maker employs crucial parameters, including online risk assessment using CPA calculations, range, and bearing to obstacles, to generate runtime decisions with clear explanations of the reasoning process. Simulation results across four fundamental maritime scenarios involving a semi-realistic ship model with manoeuvring constraints, input saturation, disturbances, and non-holonomic behaviour, demonstrate the effectiveness of our approach.

The unique capability of the algorithm lies in its ability to spell out the COLREG rules, making the decision-making process explainable. The results lead to several important conclusions. First, LLMs can interpret maritime navigation rules and generate COLREGs-compliant decisions with human-like reasoning. Second, the integration of online risk assessment with LLM decision-making enhances situational awareness and transparency of the decision-making process. Third, our risk-aware approach ensures consistent behaviour throughout encounters, preventing oscillatory decisions while maintaining safety. Advanced versions of these algorithms could even serve as captain-assist features within monitoring tools. The exponential progress in the reasoning capabilities of LLMs will further increase the likelihood of deploying such algorithms in these systems. While our fuzzy logic risk assessment model effectively combines navigation parameters into a unified risk metric, but has limitations. It may not capture all collision risk factors such as weather, vessel manoeuvreability, and traffic density, while they could be incorporated into the thresholds as varying parameters. 

Future studies should focus on testing and verifying the algorithm to build greater trust in this solution. The model currently lacks consideration of important factors like weather, vessel manoeuvrability and traffic density, which presents opportunities for enhancement in subsequent research.

\bibliographystyle{IEEEtran}
\bibliography{main}

\end{document}